# A Novel Dataset for Flood Detection Robust to Seasonal Changes in Satellite Imagery

Youngsun Jang*
South Dakota State University
Brookings, South Dakota, USA
youngsun.jang@sdstate.edu

Dongyoun Kim*
South Dakota State University
Brookings, South Dakota, USA
dongyoun.kim@jacks.sdstate.edu

Chulwoo Pack
South Dakota State University
Brookings, SD, USA
chulwoo.pack@sdstate.edu

Kwanghee Won†
South Dakota State University
Brookings, SD, USA
kwanghee.won@sdstate.edu

## ABSTRACT
This study introduces a novel dataset for segmenting flooded areas in satellite images. After reviewing 77 existing benchmarks utilizing satellite imagery, we identified a shortage of suitable datasets for this specific task. To fill this gap, we collected satellite imagery of the 2019 Midwestern USA floods from Planet Explorer by Planet Labs (Image © 2024 Planet Labs PBC). The dataset consists of 10 satellite images per location, each containing both flooded and non-flooded areas. We selected ten locations from each of the five states: Iowa, Kansas, Montana, Nebraska, and South Dakota. The dataset ensures uniform resolution and resizing during data processing. For evaluating semantic segmentation performance, we tested state-of-the-art models in computer vision and remote sensing on our dataset. Additionally, we conducted an ablation study varying window sizes to capture temporal characteristics. Overall, the models demonstrated modest results, suggesting a requirement for future multimodal and temporal learning strategies. The dataset will be publicly available on https://github.com/youngsunjang/SDSU_MidWest_Flood_2019.

## 1 INTRODUCTION
Accelerated climate change has led to more frequent extreme weather events, such as intense rainfall, snowstorms, and droughts worldwide. The catastrophic flooding in the Midwest USA in 2019 exemplifies this trend. From March to September 2019, the region experienced rapid snowmelt and flooding, significantly increasing the Mississippi River's flow and displacing about 14 million people. The floods occurred in two waves: an initial flood in March due to rapid snowmelt and a second flood in September exacerbated by reduced soil retention and heavy rainfall [6, 9]. This work focuses on the semantic segmentation of the flooded area using this event as a case study.

In remote sensing, the rise in climate change-related disasters has heightened interest in disaster detection using satellite imagery. Researchers aim to improve damage detection efficacy through Geographic Information System (GIS) expertise and advanced computing technologies. Despite the availability of semantic segmentation benchmarks like iSAID (Instance Segmentation in Aerial Images Dataset) [26] and INRIA (Institut national de recherche en sciences et technologies du numérique) [12], there is a lack of datasets specifically addressing flood occurrence. The primary open flood-related dataset, Spacenet 8 [11], focuses on object detection of buildings and roads rather than damaged area segmentation, highlighting the need for a well-prepared flood dataset.

This work addresses this gap by developing a high-resolution flood dataset using satellite images. It then applies this dataset to state-of-the-art (SOTA) computer vision models for semantic segmentation. Although these models are not specifically designed for flooded area segmentation, this study aims to report their performance for this task, providing baseline models for future research.

This work aims to contribute to the field in the following ways:
- It offers a high-quality flood dataset based on satellite images, aiding in time-series flood detection and model development resilient to seasonal changes.
- It provides experimental results on SOTA computer vision models, offering insights for future model advancements.

The remainder of this paper is organized as follows. Section 2 reviews existing benchmark datasets that utilize satellite imagery and Convolutional Neural Network (CNN)-based segmentation approaches. Section 3 details the dataset development process and introduces the core architectures of the five major segmentation models. Section 4 presents the experimental results of these baseline models, along with an ablation study under various conditions.

## 2 RELATED WORK
### 2.1 Existing Benchmark Dataset
With the recent proliferation of drone technology beyond traditional aircraft and helicopters, aerial imagery datasets are gradually expanding alongside satellite images. Since it can be challenging to differentiate between satellite imagery and aerial photographs when zooming in on localized areas, this work focuses on satellite images with resolutions tailored to the scale of towns or cities. Videos captured by drones or other aircraft are out of the scope of this study.

---
*Authors contributed equally to this work (Co-first author)
†Corresponding author

The existing 77 satellite benchamrk datasets for flood detection can be classified as shown in Table 1. This table utilizes information from https://paperswithcode.com/, an open database that provides related leaderboards [4]. Regarding non-flooding natural disasters, datasets cover various events such as volcanic eruptions [1], wildfires [2, 3, 25], landslides [18, 28], and sea ice melting [7]. Non-natural disaster datasets include land cover classification, object segmentation and classification, object tracking, scene classification, and scene generation.

The dataset developed in this study occupies a unique place among existing benchmark datasets. Details are introduced in Section 3.

## 2.2 Satellite Imagery Segmentation Models

The segmentation of flood-damaged areas in satellite images is a primary focus of study in remote sensing. Experts in subfields such as change detection often utilize GIS with emerging deep learning technologies to enhance segmentation performance. This research applies image segmentation models to address a range of downstream tasks in remote sensing.

Among the noteworthy models in this domain is the attention residual U-Net (AttResUNet) proposed by Ouyang and Li [19], built upon the traditional U-Net, a renowned CNN model for image segmentation tasks in computer vision. The U-Net is designed to learn local image features through a series of downsampling (encoder or contracting path) and upsampling (decoder or expanding path) processes centered around a bottleneck for extracting global features [22]. In this framework, skip connections preserve original visual features during upsampling, enhancing performance. AttResUNet further innovates by incorporating residual blocks of ResNet into the original U-Net architecture [10]. Additionally, AttResUNet introduces attention gates, which manage the fusion of downsampled and upsampled features using the ConvNet within an 'attention block,' diverging from conventional self-attention mechanisms.

While studies focusing on flood detection in computer vision are rare, there is a wealth of research utilizing CNN models for various object detection and semantic segmentation tasks using satellite imagery. Three prominent models in this domain include SegNeXt [8], SDSC-UNet [27], and UANet [14]. Apart from SegNeXt, SDSC-UNet and UANet commonly utilize U-Net as their backbone architecture. According to the Paperswithcode Leaderboard [12, 13], SegNeXt currently holds the top position in the semantic segmentation task on the renowned satellite benchmark iSAID (with a mIoU of 70.3) as of June 2024. SDSC-UNet and UANet are respectively ranked #2 (with a mIoU of 83.01) and #1 (with a mIoU of 83.34) on the INRIA benchmark dataset. While SegNeXt does not directly import the U-Net architecture, it shares a similar high-level architecture with U-Net, featuring an encoder-decoder structure. Its main distinction lies in its use of traditional convolution instead of transformers' self-attention to construct the encoder, known as convolutional attention, marking a departure from its prior approaches.

SDSC-UNet advances extracting buildings in satellite images, with a focus on mitigating the 'internal multiscale information' issue that previous models like ViT [5] and Swin Transformer [17] struggled with. The varying sizes of different buildings in an image pose a challenge for conventional models in effectively segmenting these multiscale objects. The innovation of SDSC-UNet lies in the incorporation of dual skip connections. These connections not only link the output of the entire transformer block of the encoder into the decoder but also integrate the attention feature maps generated at each step during the downsampling process to the decoder. This enhancement supplements the existing ViT-based models, enabling more robust segmentation of multiscale objects in satellite imagery.

UANet also aims to enhance building extraction performance using the datasets employed by SDSC-UNet. It focuses on effectively utilizing the so-called 'uncertain' feature maps during decoding. This uncertainty primarily arises from less salient buildings or complex background distributions. UANet initially employs existing models such as Feature Pyramid Networks (FPN) [15] to automatically evaluate the level of uncertainty in the image and subsequently extracts uncertain feature maps ($M_5$). Next, it conducts self-attention between the highest-level image feature map ($F_5^i$) and the $M_5$. During this self-attention process, $F_5$ in each channel dimension is involved in attention with Query and Value parameters, while $M_5$ is involved in attention with the Key parameter. The attention output is fused with $F_5$ again and concatenated with the subsequent-level feature map, thereby augmenting the model output predictions. This iterative process is repeated across all levels, leading to enhanced model performance.

While these models have shown success in tasks like object detection and semantic segmentation, their effectiveness in flood detection has not been thoroughly explored. Importantly, there are inherent differences between these tasks and flood detection, mainly due to the nature of the datasets employed. Our investigation into their performance in flood detection aims to broaden their applicability across various domains.

## 3 DATASET AND METHOD

### 3.1 Building Dataset

As mentioned in the related work, the existing 77 benchmarks that use satellite imagery are classified into three primary categories (Cat.1) and 13 subcategories (Cat.2). Category 1 delineates datasets based on their treatment of natural disasters as unusual events and whether they include or exclude relevant datasets. Category 2 further refines this classification by focusing on specific research topics within each main category.

Table 1: Benchmark Datasets Utilizing Satellite Imagery

| Cat.1 | Cat.2 | Name | Description | Source |
|---|---|---|---|---|
| Non-Natural Disaster | Land cover classification | EuroSAT | 10 classes | Sentinel-2 sat. |
| | | GID | 5 categories | Gaofen-2 sat. |
| | | PASTIS, PASTIS-R | Agricultural parcels | Sentinel-2 sat. |
| | | Five-Billion-Pixels | 24 categories | Gaofen-2 sat. |
| | | Urban Environments | 20 land use classes | Google Map |
| | | Satellite | 4 land types | WorldStrat |
| | | OSCD | 13-band, multispectral | Sentinel-2 sat. |
| | | WorldStrat | Human settlements, Time-series | Airbus SPOT 6/7 sat. Sentinel-2 sat. |
| | | SSL4EO-S12 | Time-series (Multispectral), SAR | Sentinel-1 sat. Sentinel-2 sat. |
| | | Botswana | 14 classes | NASA EO-1 sat. |
| | | HYPSO-1 Sea-Land-Cloud-Labeled Dataset | Sea-Land-Clouds | HYPSO-1 sat. |
| | | Satimage | 7 classes | |
| | Object segmentation, classification, & detection | iSAID | 15 instance categories | Google Earth, Jilin-1 sat., Gaofen-2 sat. |
| | | RoadTracer | Roads | Google Map, OpenStreetMap |
| | | SpaceNet 1 | Buildings | Worldview-2 sat. |
| | | SpaceNet 2 | Buildings | Worldview-3 sat. |
| | | SpaceNet 7 (MUDS) | Time-series, Buildings | Planet Labs, Dove |
| | | CalCROP21 | Crops | Google Earth |
| | | SICKLE | 21 crop types | Landsat-8 sat., Sentinel-1 sat., Sentinel-2 sat. |
| | | OmniCity | Buildings | Google Earth, OpenStreetMap, NYC PLUTO |
| | | fMoW | 63 categories including buildings and land use | QuickBird-2, GeoEye-1 sat., WorldView-2 &-3 sat. |
| | | xView3-SAR | Maritime objects ('dark vessels') | Sentinel-1 sat. |
| | | MASATI | 7 classes | MS Bing Map |
| | | HRPlanesV2 | Aircrafts | Google Earth |
| | | CloudCast | 10 cloud types | EUMETSAT |
| | | WHU Building Dataset | Buildings | QuickBird, Worldview, Ikonos sat., Ziyuan 3-01 sat. |
| | | iFLYTEK | Cultivated land segmentation | Jilin-1 sat. |
| | | ETDII Dataset | Electric transmission and distribution infrastructure | Aerial, WorldView-3 sat., WorldView-2 sat. |
| | | RarePlanes | Aircrafts classification | Maxar WorldView-3 sat. |
| | | University-1652 | Buildings (colleges) | Google Map, Google Earth, *Drones |
| | | BreizhCrops | Time-series | Sentinel-2 sat. |
| | | SaRNet | Missing paraglider wing | Planet Labs, Airbus, Maxar |
| | | VIGOR | Street geolocalization | Google Map |
| | | BrazilDam | Dams | Landsat 8 & Sentinel 2 sat. |
| | | MARIDA | Marine Debris | Sentinel 2 sat. |
| | | Open Buildings | Buildings | Maxar, Airbus |
| | | ELAI-Dust Storm | Dust storm | MODIS (Terra, Aqua) |
| | | RWanda Built-up Region Segmentation | Buildings | |
| | | EuroCrops | Crop types | Sentinel-2 sat. |
| | | LNDST | Water areas | Landsat 8 sat. |
| | | S2Looking | Buildings | GaoFen sat., SuperView sat., BeiJing-2 sat. |

Table 1: Benchmark Datasets Utilizing Satellite Imagery (Continued)

| Cat.1 | Cat.2 | Name | Description | Source |
|---|---|---|---|---|
| Non-Natural Disaster | | INRIA Aerial Image Labeling | Buildings | USGS National Map Service, WMS ArcGIS, Tyrol/Austria |
| | | PROBA-V | Vegetation growth | ESA PROBA-V sat. |
| | | CAESAR-Radi | Ships | Gaofen-3 sat., Sentinel-1 sat. |
| | | Sentinel 2 manually extracted deep water spectra with high noise levels and sunglint | Ocean with clouds and sunlight | Sentinel-2 1C sat. |
| | | DOTA | 15 common categories (v1.0), 18 common categories (v2.0) | Google Earth, Gaofen-2 sat. |
| | Land cover classification & Object detection | DeepGlobe 2018 | Buildings, Roads, Land cover | DigitalGlobe, SpaceNet, Vivid+ |
| | | DigitalGlobe | Buildings, Land cover | WorldView-2 & -3 sat., GeoEye-1 sat. |
| | | ShipRSImageNet | Ships | WorldView-3 sat. |
| | Obj. tracking | VISO | Moving objects (flights, cars etc.) | Jilin-1 sat. |
| | Image (scene) classification | MLRSNet | 46 categorical scenes | Google Earth |
| | | PolSF | SAR image classification | SF-ALOS2, SF-GF3, SF-RISAT, SF-RS2, SF-AIRSAR sat. |
| | | WHU-RS19 | 19 classes | Google Earth |
| | | S2-100K | Scene geolocalization | Sentinel-2 sat. |
| | | RSSCN7 | 7 classes | Google Earth |
| | Sequential generation | SCMD2016 | ConvLSTM, Cloudage | Fengyun 2-07 sat. |
| | | Weather4cast 2022 | Weather prediction | EUMETSAT, OPERA radar |
| | | EarthNet2021 | Weather prediction | Sentinel-2 sat. |
| | Paired scene generation | SEN12MS-CR | GAN, Cloud removal | Sentinel-1, & Sentinel-2 sat. |
| | | OLI2MSI | GAN, LR to HR | Landsat-8, & Sentinel-2 sat. |
| | | SEN2VEN$\mu$S | GAN, LR to HR | Sentinel-2 sat., e VEN$\mu$S sat. |
| | | WorldView-2 PairMax | Pansharpening, LR to HR (Miami) | WorldView-2 sat. |
| | | GeoEye-1 PairMax | Pansharpening, LR to HR (London & Trenton) | GeoEye-1 sat. |
| | | WorldView-3 PairMax | Pansharpening, LR to HR (Munich) | WorldView-3 sat. |
| | | Alsat-2B | Pansharpening, LR to HR (Munich) | Algeria Satellite-2B & -2A sat. |
| | | L1BSR | LR to HR | Sentinel-2 L1B sat. |
| Natural Disaster (Non-Flooding) | Volcanic areas segmentation | Hephaestus | Volcanic unrest | Sentinel-1 sat. |
| | Wildfire areas segmentation | Burned Area Delineation from Satellite Imagery | Wildfires | Sentinel-2 L2A sat. |
| | | CaBuAr | Wildfires | Sentinel-2 L2A sat. |
| | | ChaBuD | Wildfires | Sentinel-2 sat. |
| | Landslide segmentation | HR-GLDD | Landslide | PlanetScope |
| | | GVLM | Paired imagery for Landslide detection | Google Earth |
| | Glacier segmentation | CaFFe | Time-series, Glacier boundary | Sentinel-1, TerraSAR-X sat., TanDEM-X, ENVISAT sat., European RS Satellite 1&2 sat., ALOS PALSAR, RADARSAT-1 sat. |
| Natural Disaster (Flooding) | Flooded object detection | SpaceNet 8 | Flooded roads & buildings | Maxar Earth observation sat. |
| | | xBD (xView2) | Building damage (19 disaster types including flood) | Maxar/DigitalGlobe Open Data Program |
| | Flooded area segmentation | Our Dataset | Flooded land cover | Planet Labs |

Table 2: Dataset Composition. Each AOI includes one image for each of the two seasons, along with two sequential imagery of the area after the flood event.

| KS_8 Atchison county, KS (LT 39.54N, 95.08W, RB 39.52N, 95.05W) | | |
|---|---|---|
| 1-Flood | flood-1-KS_8 | May. 30, 2019, 3m/px, 3.00 m/px |
| 2-Flood | flood-2-KS_8 | Mar. 26, 2019 |
| 3-Normal | normal-1-KS_8 | Feb. 1, 2019 |
| 4-Normal | normal-2-KS_8 | Oct. 22, 2018 |
| 5-Normal | normal-3-KS_8 | Aug. 18, 2018 |
| 6-Normal | normal-4-KS_8 | Jun. 5, 2018 |
| 7-Normal | normal-5-KS_8 | Jan. 19, 2018 |
| 8-Normal | normal-6-KS_8 | Oct. 18, 2017 |
| 9-Normal | normal-7-KS_8 | Aug. 19, 2017 |
| 10-Normal | normal-8-KS_8 | May. 26, 2017 |

For instance, within the domain of flood detection, datasets are classified into object classification (damaged buildings and roads) and semantic segmentation (entire damaged areas). Consequently, the dataset developed in this study occupies a unique niche distinct from the existing benchmarks, with a primary focus on flooded area segmentation, as elucidated in Table 1.

Our dataset obtained raw satellite images from the Planet Explore service [20] (Image © 2024 Planet Labs PBC). Due to limited download availability and a supportive policy, we acquired images using screen captures. Additionally, we developed a dedicated capture program to maintain a fixed zoom level of 3.00m/px resolution on the screen. Initially, the resolution of the captured image was 810*750 pixels, which we resized to 700*700 pixels. We saved all image files in .png format and created geojson files containing coordinate information for the Area of Interest (AoI) to ensure uniform resolution across all images. Regarding binary mask images, we annotated the flooded areas using the open-source annotation tool 'Makesense' [23]. With a long-term goal of creating a model insensitive to seasonal changes, this dataset features a multi-temporal characteristic. It revisits each location 10 times at specific intervals to detect floods as unusual events.

### 3.2 Dataset Composition and Characteristics

This dataset focuses on 10 distinct locations in 5 states (Iowa, Kansas, Montana, Nebraska, and South Dakota) affected by the 2019 floods. Each location comprises 10 revisit images, encompassing 8 non-flood and 2 flood images (Figure 1). It ranges from May 2017 to October 2020, with one image per season of each year to capture seasonal variations. For instance, in Atchison County, Kansas, images were taken in different seasons starting from May 2017, with 2 additional flood images captured after the 2019 flood event (Table 2, Figure 1b). This approach helps ensure that the model does not misclassify snow-covered land in winter as flooded areas. The dataset consists of 500 images (5 states * 10 locations * 10 images), including 400 non-flood and 100 flood images.

Flooded areas are marked in white in the binary masks. This approach allowed us to create a binary mask dataset that matches the size of the images (700, 700), as shown in (Figure 2). Consequently, the dataset comprises binary mask images, with the non-flood and flood images accompanied by geojson files extracted from Planet Labs.

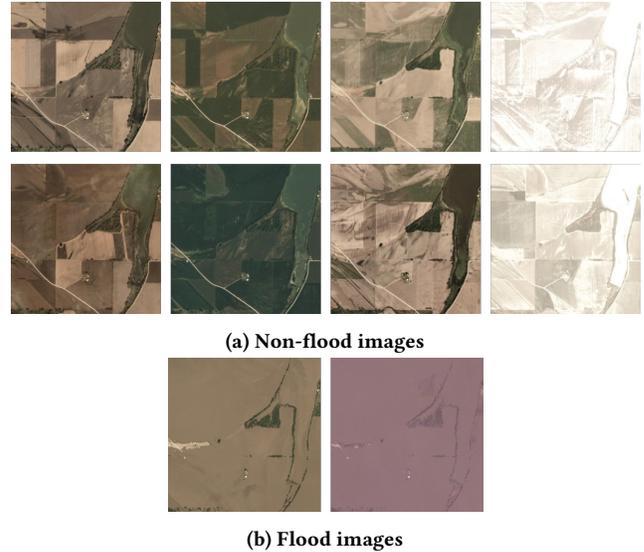

(a) Non-flood images

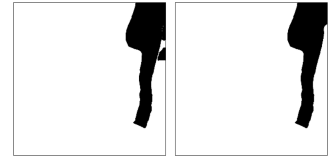

(b) Flood images

Figure 1: Multi-temporal non-flood and flood images. The temporal order proceeds from the top left to right and then from the bottom left to right (a) and left to right (b). As seen in the non-flood images, the further to the right, the more winter is depicted, with increasing snow-covered areas. From left to right, the images represent spring, summer, fall, and winter, respectively.

Figure 2: Binary masks for flooded areas. The white color denotes the flooded area, which is the focus for model training, while the black represents the non-flooded area.

We excluded pre-existing water bodies such as rivers or lakes present before the flooding event because they are not typically considered flooded areas. This decision imposes stricter conditions on model training.

In summary, we included seasonally annotated normal images to avoid misinterpreting seasonal changes as abnormal events. The dataset contains at least one image from each of the four seasons. We specifically masked only the flooded areas after the flood events, enabling the model to learn from these instances.

### 3.3 Semantic Segmentation for Flooded Areas

Semantic segmentation for each flood image requires learning the distinctions from non-flood images. We provide the model with comprehensive information about these distinctions by employing non-flood and flood pairs as model input. For instance, combining a non-flood image with its corresponding flood image channel-wise provides more reference data to the model compared to training with flood images alone. This approach also helps augment the limited flood dataset, which consists of only 100 images.

Unlike the change detection task in remote sensing, where a model is trained solely on individual non-flood and flood images, this approach expands the training dataset by generating multiple combinations using the 8 non-flood images and 2 flood images available for each location.

To enhance the dataset, we use sliding windows in the pairing process. For example, it can create 16 non-flood and flood image pairs with a window size of $w$=1 at a location. These pairs range from *(normal_1, flood_1)* to *(normal_8, flood_1)* and from *(normal_1, flood_2)* to *(normal_8, flood_2)*. Increasing the window size to $w$=2 extends the pairs to include *(normal_1, normal_2, flood_1)*, *(normal_2, normal_3, flood_1)*, and so on. This pattern continues with increasing window size, facilitating more comprehensive training data generation.

We applied the experimental conditions to five baselines, and the ablation study in the subsequent section presents detailed results of the experiments.

## 4 EXPERIMENTAL RESULTS

The five baseline models for the experiment are the original U-Net, AttResUNet, SegNeXt, SDSC-UNet, and UANet. These are primary CNN models for semantic segmentation in computer vision. Image pair sets and mask data were employed to evaluate the segmentation performance of these models. The evaluation metrics include accuracy, Dice coefficient, and mean Intersection over Union (mIoU). Accuracy measures the actual predictions among the total predictions (5), while the Dice coefficient and mIoU focus on the foreground.

$$Accuracy = \frac{TP + TN}{TP + TN + FN + FP} \quad (1)$$

The difference between these metrics lies in that while Dice uses the total area with overlapping between prediction and ground truth as the denominator (6), IoU uses the union without overlapping between prediction and ground truth as the denominator (7). These metrics are more widely used for datasets when classes are imbalanced.

$$Dice = \frac{2 * TP}{2 * TP + FP + FN} \quad (2)$$

$$IoU = \frac{TP}{TP + FP + FN} \quad (3)$$

The experiment results in Table 3 show that the overall accuracy of the models, except SDSC-UNet, did not surpass 80%, and the mIoU did not reach the levels achieved by those models in the iSAID and INRIA datasets. It indicates the necessity for performance enhancement through fine-tuning or more advanced approaches, such as multimodal learning.

In addition, Table 3 reveals notable insights from repeated experiments. The original U-Net consistently showed the lowest performance, whereas SDSC-UNet consistently outperformed all other models across all metrics. However, the performance of the remaining three models was largely similar, underscoring the substantial influence of data variability on the outcomes.

Table 3: Best Performance of 5 Baseline Models

|  | Val loss | Accuracy | Dice | mIoU |
| --- | --- | --- | --- | --- |
| Original U-Net | 0.7236 | 0.5054 | 0.2618 | 0.1742 |
| AttResUNet | 0.4860 | 0.7577 | 0.5912 | 0.5098 |
| SegNeXt | 0.6328 | 0.6798 | 0.7386 | 0.6300 |
| **SDSC-UNet** | **0.4058** | **0.8437** | **0.8331** | **0.7543** |
| UANet | 2.0640 | 0.7717 | 0.6334 | 0.5240 |

### 4.1 Ablation Study on the Window Size $w$

Table 4 displays the experimental results for each window size $w$. This study found no significant differences in performance attributable to the window size. Instead, we observed that performance differences were more likely influenced by the model architecture. Once again, SDSC-UNet consistently demonstrated higher performance compared to the other models.

These findings suggest two important points. Firstly, SDSC-UNet's architecture appears relatively well-suited for segmenting flooded areas in our dataset. However, its experimental results varied significantly across runs, necessitating cautious interpretation and validation through additional experiments. It appears that all models are significantly affected by the randomness inherent in the input data. Secondly, applying window sizes may not adequately capture the multi-temporal characteristics of our dataset. Concatenating sequential images alone may be not adequate for extracting temporal features; thus, employing models specifically designed for learning from time-series satellite imagery data seems necessary.

## 5 CONCLUSION

This study explored the need for a novel dataset designed to detect uncommon events such as floods in satellite imagery. It reviewed existing benchmark datasets and introduced our new dataset, detailing its unique characteristics and composition. Additionally, the study evaluated the performance of SOTA models in computer vision and remote sensing using this dataset. Due to the overall limited performance of the models, an analysis of relative performance differences was excluded from this study.

The experimental results generally demonstrated modest performance, highlighting the challenges posed by our proposed dataset. We attribute this difficulty to the complexity of interpreting multi-temporal features inherent in the dataset, as current SOTA models primarily rely on visual features alone. To address this gap in future research, we intend to develop an advanced flooded areas segmentation model that integrates multimodal approaches to effectively incorporate these multi-temporal features during the learning process.

Specifically, our plans include exploring more sophisticated multimodal approaches. Building on the concept of RemoteCLIP [16], we aim to fine-tune the CLIP model [21] and conduct an ablation study to determine the optimal method for integrating extracted image and geographical text features into an optimized multimodal setup. This iterative process will guide us in selecting the most appropriate model configurations. In addition, we will investigate methods designed to effectively manage time-series data. For instance, ViTs for Satellite Image Time Series (SITS) presents a promising model for analyzing time-series satellite imagery [24], offering an opportunity to exploit the dataset's intricate temporal attributes.

Table 4: Performance by Different Window Size

|  | Val loss | Accuracy | Dice | mIoU |
|---|---|---|---|---|
| Original U-Net |  |  |  |  |
| $w=1$ | 0.7031 | 0.5209 | 0.1834 | 0.1288 |
| $w=2$ | 0.6838 | 0.5554 | 0.1921 | 0.1390 |
| $w=3$ | 0.7103 | 0.4997 | 0.1329 | 0.0852 |
| $w=4$ | 0.7091 | 0.5032 | 0.1472 | 0.0868 |
| $w=5$ | 0.7042 | 0.5136 | 0.1706 | 0.1115 |
| $w=6$ | 0.6864 | 0.5531 | 0.2067 | 0.1485 |
| $w=7$ | 0.6954 | 0.5241 | 0.1436 | 0.0936 |
| $w=8$ | 0.7236 | 0.5054 | 0.2618 | 0.1742 |
| AttResUNet |  |  |  |  |
| $w=1$ | 0.6258 | 0.5321 | 0.0171 | 0.0092 |
| $w=2$ | 0.6355 | 0.6308 | 0.3528 | 0.2601 |
| $w=3$ | 0.4860 | 0.7577 | 0.5912 | 0.5098 |
| $w=4$ | 0.5638 | 0.7305 | 0.5106 | 0.4206 |
| $w=5$ | 0.5810 | 0.6224 | 0.3708 | 0.2736 |
| $w=6$ | 0.5294 | 0.7138 | 0.2293 | 0.1670 |
| $w=7$ | 0.6544 | 0.5603 | 0.1569 | 0.1175 |
| $w=8$ | 0.6203 | 0.6980 | 0.1657 | 0.1333 |
| SegNeXt |  |  |  |  |
| $w=1$ | 0.6875 | 0.5502 | 0.6696 | 0.5502 |
| $w=2$ | 0.6629 | 0.5834 | 0.3079 | 0.2470 |
| $w=3$ | 0.6930 | 0.4813 | 0.4867 | 0.3609 |
| $w=4$ | 0.6328 | 0.6798 | 0.7386 | 0.6300 |
| $w=5$ | 0.6952 | 0.4411 | 0.3440 | 0.2367 |
| $w=6$ | 0.5765 | 0.7231 | 0.6129 | 0.5287 |
| $w=7$ | 0.6893 | 0.5702 | 0.6931 | 0.5675 |
| $w=8$ | 0.6966 | 0.4770 | 0.3346 | 0.2546 |
| SDSC-UNet |  |  |  |  |
| $w=1$ | 0.4472 | 0.8018 | 0.8117 | 0.7143 |
| $w=2$ | 0.5470 | 0.7660 | 0.7009 | 0.5882 |
| $w=3$ | 0.4468 | 0.8045 | 0.8098 | 0.7062 |
| $w=4$ | 0.5658 | 0.7406 | 0.7177 | 0.6009 |
| $w=5$ | 0.5430 | 0.7378 | 0.6904 | 0.5752 |
| $w=6$ | 0.4942 | 0.7844 | 0.8211 | 0.7161 |
| $w=7$ | 0.4058 | 0.8437 | 0.8331 | 0.7543 |
| $w=8$ | 0.4683 | 0.7758 | 0.7100 | 0.5925 |
| UANet |  |  |  |  |
| $w=1$ | 1.9917 | 0.7661 | 0.5171 | 0.4115 |
| $w=2$ | 2.3104 | 0.7256 | 0.6259 | 0.5038 |
| $w=3$ | 2.0640 | 0.7717 | 0.6334 | 0.5240 |
| $w=4$ | 2.3622 | 0.6692 | 0.6207 | 0.4904 |
| $w=5$ | 2.0850 | 0.7546 | 0.5692 | 0.4654 |
| $w=6$ | 3.0059 | 0.5231 | 0.5674 | 0.4070 |
| $w=7$ | 2.7793 | 0.5888 | 0.5344 | 0.4059 |
| $w=8$ | 2.7269 | 0.5995 | 0.3035 | 0.2227 |

## ACKNOWLEDGMENTS

This research was supported by the South Dakota NASA EPSCoR Program (NASA grant 80NSSC22M0045)